\documentclass{article} 
\usepackage{iclr2016_workshop,times}
\usepackage{hyperref}
\usepackage{url}

\usepackage[T1]{fontenc}
\usepackage[latin1]{inputenc}
\usepackage{amssymb}
\usepackage{amsmath}
\usepackage{amsfonts}
\usepackage{graphicx}
\usepackage[algo2e,noend,boxed]{algorithm2e}
\usepackage{algorithmic}
\usepackage{algorithm}

\usepackage{xspace}
\usepackage{dsfont}
\usepackage{graphicx}  
\usepackage{verbatim}
\usepackage{appendix}
\usepackage{multirow}
\usepackage{color}
\usepackage{framed}
\usepackage{wrapfig}

\definecolor{shadecolor}{rgb}{.6,.6,.6}
 {\vspace*{0.5\baselineskip}
\begin{center}\begin{minipage}{0.85\textwidth}\begin{shaded}}%
 {\end{shaded}\end{minipage}\end{center} \vspace*{0.5\baselineskip}}

\usepackage[tight,scriptsize]{subfigure}

\newcommand{\Jon}[1]{{\color{red}{\bf\sf [Jon: #1]}}}

\newcommand{\eat}[1]{}

\numberwithin{equation}{section}

\newcommand{\blockcomment}[1]{ }

\newbox\subfigbox


\include{wrapfig}
\title{Efficient inference in occlusion-aware generative models of images}

\author{Jonathan Huang \& Kevin Murphy \\
Google Research \\
1600 Amphitheatre Parkway \\
Mountain View, CA 94043, USA\\
\texttt{\{jonathanhuang, kpmurphy\}@google.com}
}

\begin{document}

\maketitle

\begin{abstract}
We present a generative model of images based on layering,
in which image layers are individually generated, then composited
from  front to back.  We are thus able to factor the appearance
of an image into the appearance of individual objects within the image --- and additionally for
each individual object, we can factor content from pose.  
Unlike prior work on layered models, we learn a shape prior for each
object/layer, allowing 
the model to tease out which object is in front by looking for a consistent shape,
without needing access to motion cues
or any labeled data.
We show that ordinary stochastic gradient variational bayes (SGVB), which optimizes our fully differentiable
lower-bound on the log-likelihood, is sufficient to learn an interpretable representation of images.
Finally we present experiments demonstrating the effectiveness of the model for inferring foreground
and background objects in images.
\end{abstract}

\section{Introduction}

\begin{wrapfigure}[11]{r}[12pt]{6cm}
\vspace{-4mm}
\begin{center}
    \includegraphics[width=0.375\textwidth]{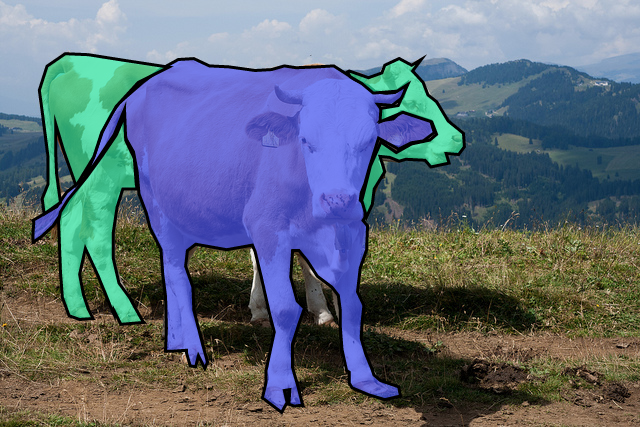} \vspace{-2mm}
    \caption{\footnotesize Illustration of occlusion.}
    \label{fig:cows}
  \end{center}
  \vspace{-2mm}
\end{wrapfigure}

Recently computer vision has made great progress by training deep
feedforward neural networks on large labeled datasets.
However, acquiring labeled training data for all of the problems that
we care about is expensive.
Furthermore, some problems require top-down inference as well as
bottom-up inference in order to handle ambiguity.
For example, consider the problem of object detection
and instance segmentation  in the
presence of clutter/occlusion.,
 as illustrated in Figure~\ref{fig:cows}.
In this case, the foreground object may
obscure almost all of the background object, yet people are still able
to detect that there are two objects present, to correctly segment
out both of them, and even to amodally complete the hidden parts of
the occluded object (cf., \cite{Kar2015}).

One way to tackle this problem is to use generative models.
In particular, we can imagine the following generative process for an image:
(1) Choose an object (or texture) of interest, by sampling a ``content vector''
representing its class label, style, etc;
(2) Choose where to
place the object in the 2d image plane, by sampling a ``pose vector'',
representing location, scale, etc. 
(3) Render an image of the object onto a hidden canvas or layer;\footnote{
We  use  ``layer''  in this paper mainly to refer to image layers, however
in the evaluation section (Section~\ref{sec:eval}) ``layer'' will also be used to refer to
neural network layers where the meaning will be clear from context.
}
(4) Repeat this process for $N$
objects (we assume in this work that $N$ is fixed);
(5) Finally, generate
the observed image by compositing the layers in order.\footnote{
There are many  ways to composite multiple layers in computer
graphics~\citep{porter1984compositing}. 
In our experiments, we use  the classic \emph{over operator}, which
reduces to a simple $\alpha$-weighted 
convex combination of foreground and background pixels,
in the two-layer setting.
 See Section~\ref{sec:models} for more details.
}

\looseness -1 There have been several previous attempts to use layered generative
models to perform scene parsing and object detection in clutter (see
Section~\ref{sec:related} for a review of related work). However, such
methods usually run into computational bottlenecks, since inverting
such generative models is intractable.
In this paper, we build on recent work (primarily \cite{Kingma2014,gregor2015draw})
that shows how to jointly train a generative model and an inference
network in a way that optimizes a variational lower bound on the log
likelihood of the data;
this has been called a ``variational auto-encoder'' or VAE.
In particular, we extend this prior work in
two ways. First, we extend it to the sequential setting, where we
generate the observed image in stages by compositing hidden layers
from front to back. 
Second, we combine the VAE with the spatial transformer network of
\citep{jaderberg2015spatial}, allowing us to factor out variations
in pose (e.g., location) from variations in content (e.g., identity).
We call our model the ``composited spatially transformed VAE'', or CST-VAE
for short.

Our resulting inference algorithm combines top-down (generative)
and bottom-up (discriminative) components in an interleaved fashion as follows:
(1)  First we recognize (bottom-up) the foreground object, factoring apart pose and content;
(2) Having recognized it, we generate (top-down) what the hidden image  should look
like;
(3) Finally, we virtually remove this generated hidden image from the observed
image to get the residual image, and we repeat the process.
(This is somewhat reminiscent of approaches that the brain is believed
to use, \cite{Hochstein2002}.)

The end result is a way to factor an observed image of
overlapping objects into $N$
hidden layers, where each layer contains a single object
with its pose parameters.
Remarkably, this whole process can be trained in a fully unsupervised
way using standard gradient-based optimization methods
(see Section~\ref{sec:models} for details).
In Section~\ref{sec:eval}, we show that our method is able to reliably
interpret cluttered images, and that the inferred latent
representation is a much better  feature vector for a discriminative
classification task than working with the original cluttered images.

\eat{
Large strides have been made in computer vision in recent years 
due to scalable deep learning and massive labeled datasets such as Imagenet~\cite{krizhevsky2012imagenet,deng2009imagenet}.
But as we move beyond simple image classification problems to more complication output spaces, the limitations
to our current approaches are becoming more transparent.
Dense labeling problems such as semantic segmentation, instance segmentation, for example,
require much more human effort per image to label and more images in general.

We've seen less progress on unsupervised models  ---  using MCMC

traditional graphical models from yesteryear (citations?) 	- methods tend to be less scalable often requiring intractable normalization constants to be computed
  --- recent work has bypassed this (citations) allowing models to be trained via gradient methods

Things don't just have one label

We also are starting to get interested in dense labels of images and video: think semantic segmentations, instance level segmentations, depth, etc.
	It's getting more and more expensive to collect this data.

Unsupervised and weakly supervised models are the next frontier.  

The goal is to model the variation in real images.  
Disentangling the factors of variation is critical to confronting the curse of dimensionality.

conv nets for example learn higher levels of abstraction, disentangling variation in appearance

there has been some work in the deep learning community factoring pose from appearance.

No work on factoring out the variability stemming from Multiple objects.

We propose a layered model of image generation.  And it has these benefits.

\Jon{play up interpretability of the model}

A critical issue in counting and instance level segmentation is dealing with things that can overlap and partially occlude each other, and current detection methods that use non-max suppression, for example, are not smart at reasoning about overlap/occlusion.  

In the instance segmentation problem, for example, we might need to analyze a local patch and decide whether it belongs to instance A, or instance B, or whether there were even two instances that were overlapping to begin with.  In some of these settings, the decision cannot be made at the local level and requires a global understanding of the semantics of the scene.

Motivations:
	The idea that we explore here is: if we only knew what the instances A and B were supposed to look like (i.e., had pixel-level generative models of A and B), we would be better positioned to make this decision.

	Being able to do this can help in a ton of applications: counting, instance segmentation
		even training for object recognition ? a lot of examples are occluded and if knew how
			how to factor out this variation, it could help
			
	But we can go further: we can imagine what?s behind (amodal completion)

	A step toward integrating top-down and bottom-up
	
	part of the story is that we're able to use VAEs to now combine some of the prior structure that we know about 
		with the rich expressiveness of deep models and train them all together with backprop

We propose a generative model that separately generates a ?layer? for each object in the image
in which each layer itself we?ve separated pose from style.
\begin{itemize}
\item fully unsupervised training (?) using ordinary sgd based methods that use backprop
\item st-aevb model
\item new ?alpha compositing? layer which is kind of like a pixelwise attention mechanism
\item masked inference mechanism allowing for training from partially observed images
\item usage of conv and deconv layers in a fully probabilistic generative model
\item show how to reason with occlusions and do amodal completion of objects based on a generative model
can handle multiple and unknown number of objects?
\end{itemize}
}

\section{Related work}
\label{sec:related}
\vspace{-2mm}

\subsection{Deep probabilistic generative models}
\vspace{-2mm}
Our approach is inspired by the recent introduction of generative deep learning models that can be
trained end-to-end using backpropagation.  These models have included
generative adversarial networks \citep{denton2015deep, goodfellow2014generative}
as well as variational auto-encoder (VAE) models \citep{Kingma2014,kingma2014semi,rezende2014stochastic,burda2015importance}
which are most relevant to our setting.  

Among the variational auto-encoder literature, our work is most comparable to 
the DRAW network of~\cite{gregor2015draw}.  As with our proposed model, 
the DRAW network is a generative model of images in the variational auto-encoder framework
that decomposes image formation into multiple stages of additions to a canvas matrix.  The DRAW paper assumes an LSTM based generative model of these sequential 
drawing actions which is more general than our model.
In practice, these drawing actions seem to progressively refine an initially blurry region of an image to  be sharper.
In our work, we also construct the image sequentially,
but each step is encouraged to correspond to a layer in the image,
similar to what one might have in typical photo editing software.
This encourages our hidden stochastic variable to be interpretable,
which could potentially be useful for semi-supervised learning.

\vspace{-2mm}
\subsection{Modeling Transformation in neural networks}\vspace{-2mm}
\looseness -1 One of our major contributions is a model that is capable of separating the pose of an object from its appearance,
which is of course a classic problem in computer vision.
Here we highlight several of the most related works from 
the deep learning community.  
Many of these  related works have been influenced by the Transforming Auto-encoder models by \cite{hinton2011transforming},
in which pose is explicitly separated from content in an auto-encoder which is trained to predict (known)
small transformations of an image.
More recently,  \cite{dosovitskiy2014learning} introduced a convolutional network to generate images of chairs where pose was explicitly separated out, and \cite{cheung2014discovering} introduced an auto-encoder where a subset of variables such as pose can be explicitly observed and remaining
variables are encouraged to explain orthogonal factors of variation.   
Most relevant in this line of works is that of~\cite{kulkarni2015deep}, which, like us, 
separate the content of an image from pose parameters using a variational auto-encoder.
In all of these works, however, there is an element of supervision, where variables such as pose
and lighting are known at training time. 
Our method,
which is based on the recently introduced Spatial
Transformer Networks paper
 ~\citep{jaderberg2015spatial},
is able to separate pose from content in an fully unsupervised setting 
using standard off-the-shelf gradient methods.

\eat{
Finally our method relies crucially on the recently introduced Spatial
Transformer Networks paper
 ~\citep{jaderberg2015spatial}, which introduced 
a fully differentiable module that allows one to warp/rotate images based on some input spatial transformation.
To our knowledge, we are the first to apply this module in an unsupervised learning setting.  
}

\vspace{-2mm}
\subsection{Layered models of images}\vspace{-2mm}

Layer models of images is an old idea.
Most works take advantage of motion cues to decompose video data into layers~\citep{darrell1991robust,wang1994representing,ayer1995layered,kannan2005generative,kannan2008fast}.
However, there have been some papers that work from single images.
\cite{yang2012layered}, for example, propose a layered model for
segmentation but rely heavily on bounding box and categorical
annotations.	
\cite{Isola2013} deconstruct a single image into layers, but require a
training set of manually segmented regions.
Our generative model is similar to that proposed by \cite{williams2004greedy}, however we can capture more complex
appearance models by using deep neural networks, compared to their per-pixel mixture-of-gaussian models.  Moreover,
our training procedure is simpler since it is just end-to-end minibatch SGD.
Our approach also has similarities to the work of \cite{le2011learning}, however  they use restricted Boltzmann machines, 
which require expensive MCMC sampling to estimate gradients and have difficulties reliably estimating the 
log-partition function.  
\vspace{-2mm}
	


\section{CST-VAE: A probabilistic layered model of image generation}
\label{sec:models}\vspace{-3mm}

\begin{figure}[t]
\begin{center}
\subfigure[]{
\raisebox{1mm}{
\includegraphics[width=0.30\linewidth]{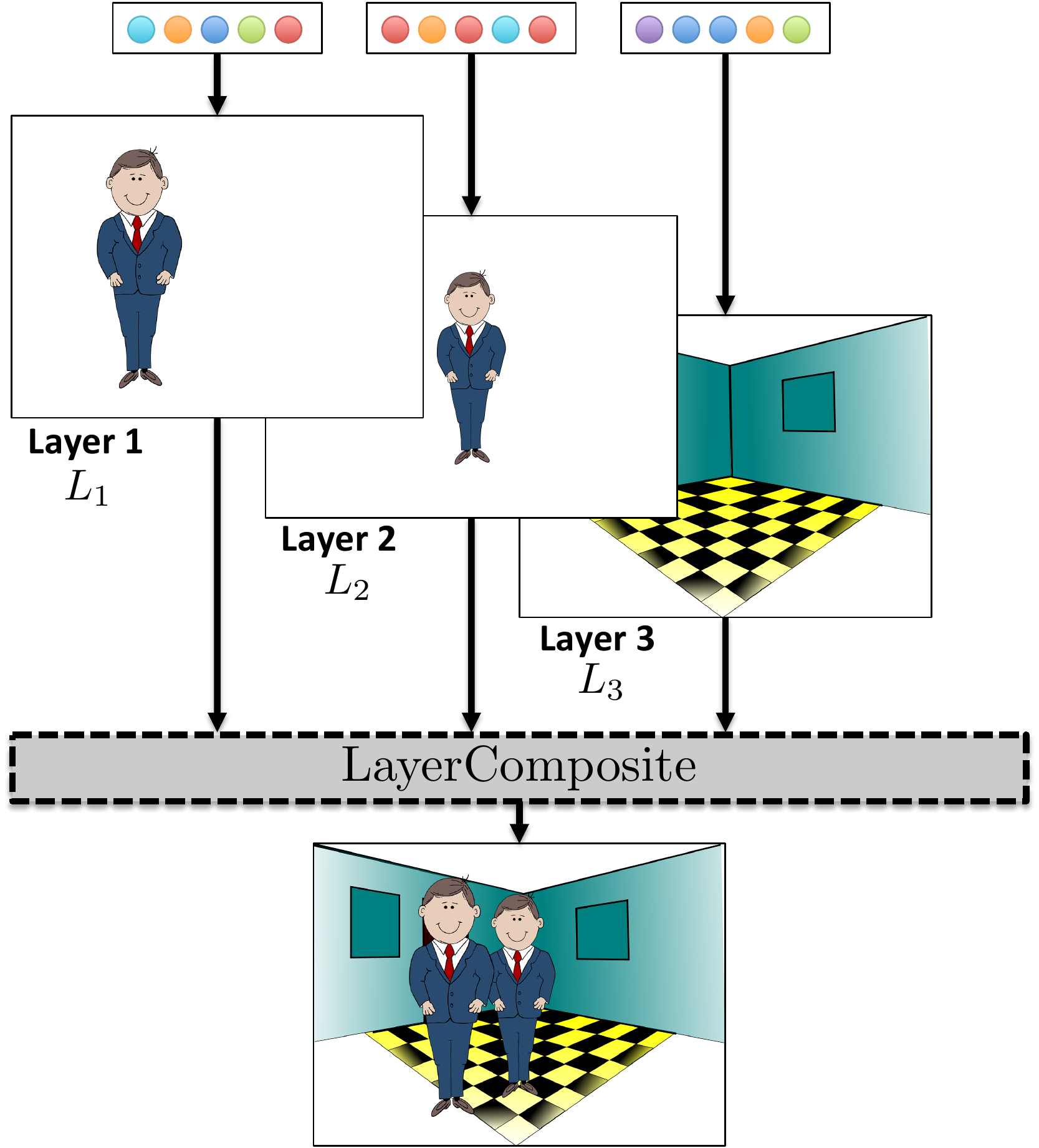}
\label{fig:cartoon}
}
}\qquad\qquad\qquad
\subfigure[]{
\includegraphics[width=0.35\linewidth]{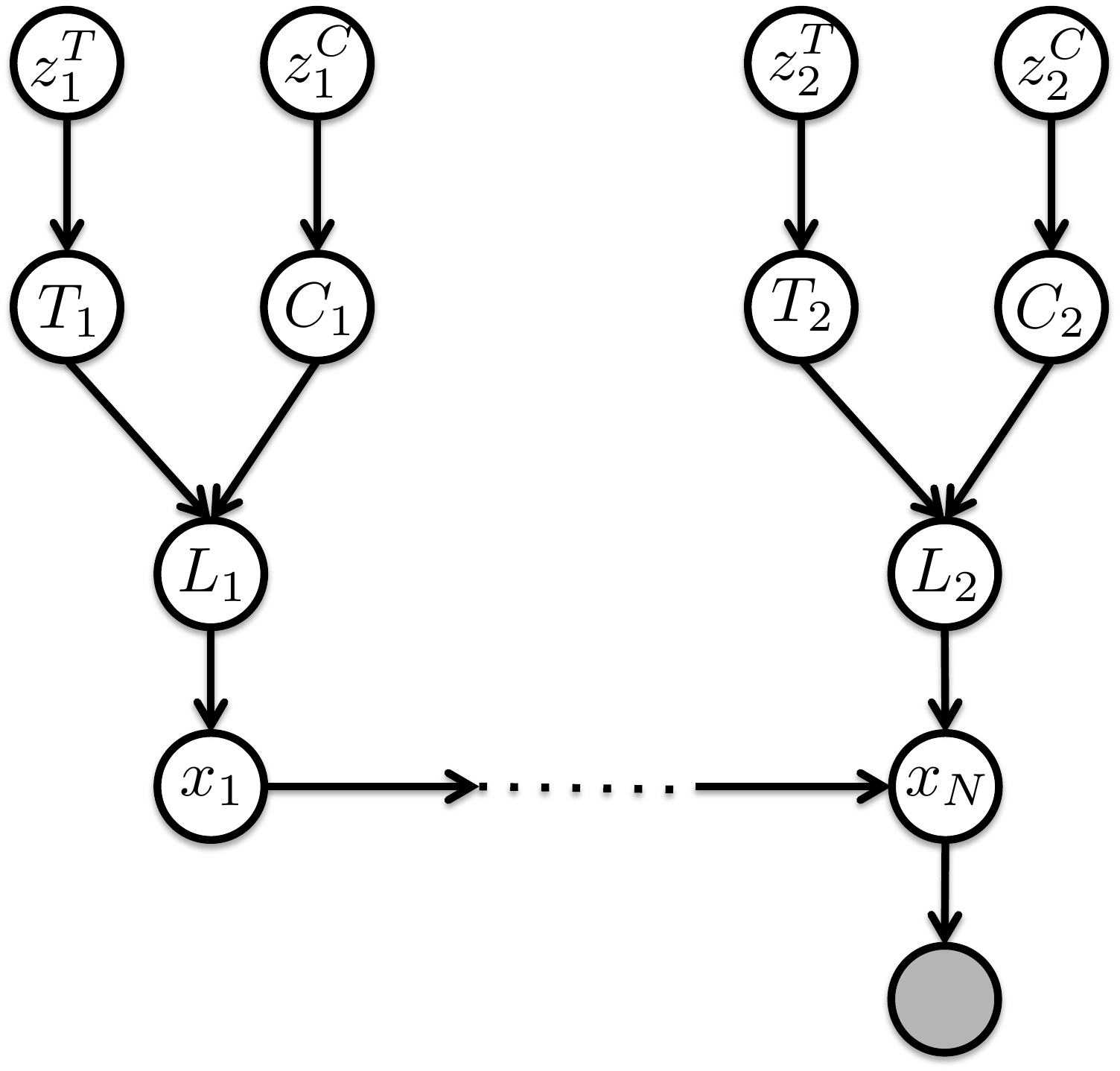}
\label{fig:compositestvae_gm}
}\vspace{-5mm}
\end{center}
 \caption{\footnotesize \subref{fig:cartoon} Cartoon illustration of the CST-VAE layer compositing process; 
 \subref{fig:compositestvae_gm} CST-VAE graphical model.  
 }\vspace{-3mm}
\end{figure}

\looseness -1 In this section we introduce the \emph{Composited Spatially Transformed Variational Auto-encoder  (CST-VAE)},
a family of latent variable models, which factors
the appearance of an image into the appearance of the different layers
that create that image.  
Among other things, the CST-VAE model allows us to tease
apart the component layers (or objects) that make up an image and
reason about occlusion in order to perform tasks such as amodal
completion 
\citep{Kar2015}
or instance segmentation
\citep{Hariharan2014}.
Furthermore, it can be trained in a fully unsupervised fashion using  minibatch stochastic
gradient descent methods but can also make use of labels in supervised or semi-supervised settings.
 
 In the CST-VAE model, we assume that images are created by (1) generating a sequence of image layers,
 then (2) compositing the layers to form a final result.  Figure~\ref{fig:cartoon} shows a simplified cartoon illustration of this process.  
 We now discuss these two steps individually.
 \vspace{-1mm}
 \paragraph{Layer generation via the ST-VAE model.}
\looseness -1 The layer generation model is interesting in its own right and we will call it the \emph{Spatially transformed 
 Variational Auto-Encoder (ST-VAE) model} (since there is no compositing step).
 We intuitively think of  layers as corresponding to objects in a scene --- a layer $L$ is assumed 
 to be generated by first generating an image $C$ of an object in some canonical pose (we refer to this image as
 the \emph{canonical image} for layer $L$), then warping $C$ in the 2d image plane (via some transformation $T$).
We assume that both $C$ and $T$ are generated by some latent variable --- specifically $C = f_C(z^C; \theta_C)$ and $T = f_T(z^T; \theta_T)$,
where $z^C$ and $z^T$ are latent variables and $f_C(\cdot; \theta_C)$ and $f_T(\cdot; \theta_T)$ are nonlinear functions with parameters
$\theta_C$ and $\theta_T$ to be learned.  We will call these content
and pose generators/decoders.
We are agnostic as to the particular parameterizations of $f_C$ and $f_T$, though as we discuss below, they
are assumed to be almost-everywhere differentiable and in practice we
have used MLPs.
In the interest of seeking simple interpretations of images, we also assume that these latent pose and content ($z^C$, $z^T$) 
variables are low-dimensional and independently Gaussian.

Finally to obtain the warped image, we
use \emph{Spatial Transformer Network} (STN) modules, recently introduced by~\cite{jaderberg2015spatial}.
We will denote  the result of resampling an image $C$ onto a regular grid which has been transformed by $T$ by $STN(C, T)$.
The benefit of using STN modules in our setting 
is that they perform resampling in a differentiable way, allowing for our models to be trained using gradient methods.
 \vspace{-1mm}
\paragraph{Compositing.}
To form the final observed image of the (general multi-layer) CST-VAE model, we generate a sequence of layers $L_1$, $L_2$, \dots, $L_N$
independently drawn from the ST-VAE model and composite from
front to back.
There are many  ways to composite multiple layers in computer graphics~\citep{porter1984compositing}.
In our experiments, we use  the classic \emph{over operator}, which reduces to a simple $\alpha$-weighted
convex combination of foreground and background pixels (denoted as a  binary  operation $\oplus$)
in the two-layer setting, but can be iteratively applied
to handle multiple layers. 

To summarize, the CST-VAE model can be written as the following generative process.  Let $x_0=\mathbf{0}^{w\times h}$ (i.e., a black image).   For $i=1,\dots,N$:\vspace{-4mm}

{\footnotesize
\begin{align*}
z_i^C, z_i^T  &\sim \mathcal{N}(0, I), \\
C_i &= f_C(z_i^C; \theta_C), \\
T_i &= f_T(z_i^T; \theta_T), \\
L_i &= \mbox{STN}(C_i, T_i), \\
x_i &= x_{i-1} \oplus L_i,
\end{align*}\vspace{-4mm}
} 

Finally, given $x_N$, we stochastically generate the observed image
$x$ using $p(x|x_N)$.
If the image is binary, this is a Bernoulli
model of the form $p(x^j=1|x_N^j) = \mbox{Ber}(\sigma(x_N^j))$
for each pixel $j$;
if the image is real-valued, we use a Gaussian model of the form
$p(x^j=1|x_N^j) = {\cal N}(x_N^j, \sigma^2)$.
See Figure~\ref{fig:compositestvae_gm} for
 a graphical model depiction of the CST-VAE generative model.

\subsection{Inference and parameter learning with variational auto-encoders}

In the context of the CST-VAE model, we are  interested in two coupled
problems: 
\textbf{inference}, by which we mean inferring all of the latent
variables $z^C_i$ and $z^T_i$ given the model parameters $\theta$
and the image;
and 
 \textbf{learning}, by which we mean estimating the model parameters
$\theta=\{\theta_C,\theta_T\}$ given a training set of images
$\{x^{(i)}\}_{i=1}^m$.
Traditionally for latent
variable models such as CST-VAE, one might solve these problems using
EM ~\citep{dempster1977maximum},
using approximate inference (e.g., loopy
belief propagation, MCMC or mean-field)
in the E-step (see e.g.,~\cite{wainwright2008graphical}).
 However if we want to allow for
rich expressive parameterizations of the generative models $f_C$ and
$f_T$, these 
approaches become intractable.  Instead we use the recently proposed
variational auto-encoder (\emph{VAE}) framework~\citep{Kingma2014} for
inference and learning.

In the variational auto-encoder setting, we assume that the posterior
distribution over latents is parameterized by a particular form 
 $Q(z^C, z^T|\gamma)$,
where $\gamma$ are data-dependent parameters.
Rather than optimizing these at runtime,
we compute them using 
an MLP, $\gamma=f_{enc}(x,\phi)$, which is called a \emph{recognition model} or an
\emph{encoder}.
We jointly optimize the generative model parameters $\theta$ and
recognition model parameters $\phi$ by maximizing the following:
\begin{equation}\label{eqn:vae_objective}\footnotesize
\mathcal{L}(\theta, \phi; \{x^{(i)}\}_{i=1}^m)
	= \sum_{i=1}^m \frac{1}{S}\sum_{s=1}^S \left[-\log
          Q(z_{i,s}^C, z_{i,s}^T | f_{enc}(x^{(i)},\phi)) 
+ \log P(x^{(i)} | z_{i,s}^C, z_{i,s}^T; \theta) \right],
\end{equation}
where $z_{i,s}^C, z_{i,s}^T \sim Q(z^C, z^T | f_{enc}(x^{(i)};\phi)) $ are
samples drawn from the variational posterior $Q$, and $m$ is the size
of the training set,
and
$S$ is the number of times we must sample the posterior per training example
(in practice, we use $S=1$, following~\cite{Kingma2014}).
We will use a diagonal multivariate Gaussian for $Q$, so that
the recognition model just has to predict the mean and  variance,
 $\mu(x;\phi)$ and $\sigma^2(x;\phi)$.

Equation~\ref{eqn:vae_objective} is stochastic lower bound on the observed data log-likelihood and interestingly,
is differentiable with respect to parameters $\theta$ and $\phi$ in certain situations. In particular,
when $Q$ is Gaussian and the likelihood under the generative model $P(x|z^C,z^T; \theta)$ is differentiable, 
then the stochastic variational lower bound can be written in an end-to-end
differentiable way via the so-called \emph{reparameterization trick} introduced in ~\cite{Kingma2014}.
Furthermore, the objective in Equation~\ref{eqn:vae_objective} can be interpreted as a reconstruction cost plus regularization term on the 
bottleneck portion
of a neural network, which is why we think of these models as auto-encoders.  
In the following, we discuss how to do parameter learning and inference for the CST-VAE model more specifically.
The critical design choice that must be made is how to parameterize  the recognition model so that we can 
appropriately capture the important dependencies that may arise in the posterior.
\vspace{-3mm}

\eat{
In the following, we discuss how to do parameter learning and inference for the CST-VAE model
within the variational auto-encoder framework more specifically.
The critical design choice that must be made is how to parameterize  the recognition model $Q$. There are two things 
that we hope for in a recognition model: (1) it must be able to 
capture the important dependencies that may arise in the posterior; (2) we should be able to differentiate the result of
sampling $Q$ with respect to its parameters $\phi$.  The second desiderata is accomplished (as is standard in VAE models)
by parameterizing $Q$ as a normal distribution (whose mean and variance are nonlinear functions of $\phi$), allowing us to differentiate using the 
reparametrization trick.  
}

\subsection{Inference in the ST-VAE model}
\label{sec:stvae}\vspace{-3mm} 

\begin{figure}[t]
\begin{center}
\subfigure[]{
\includegraphics[width=0.25\linewidth]{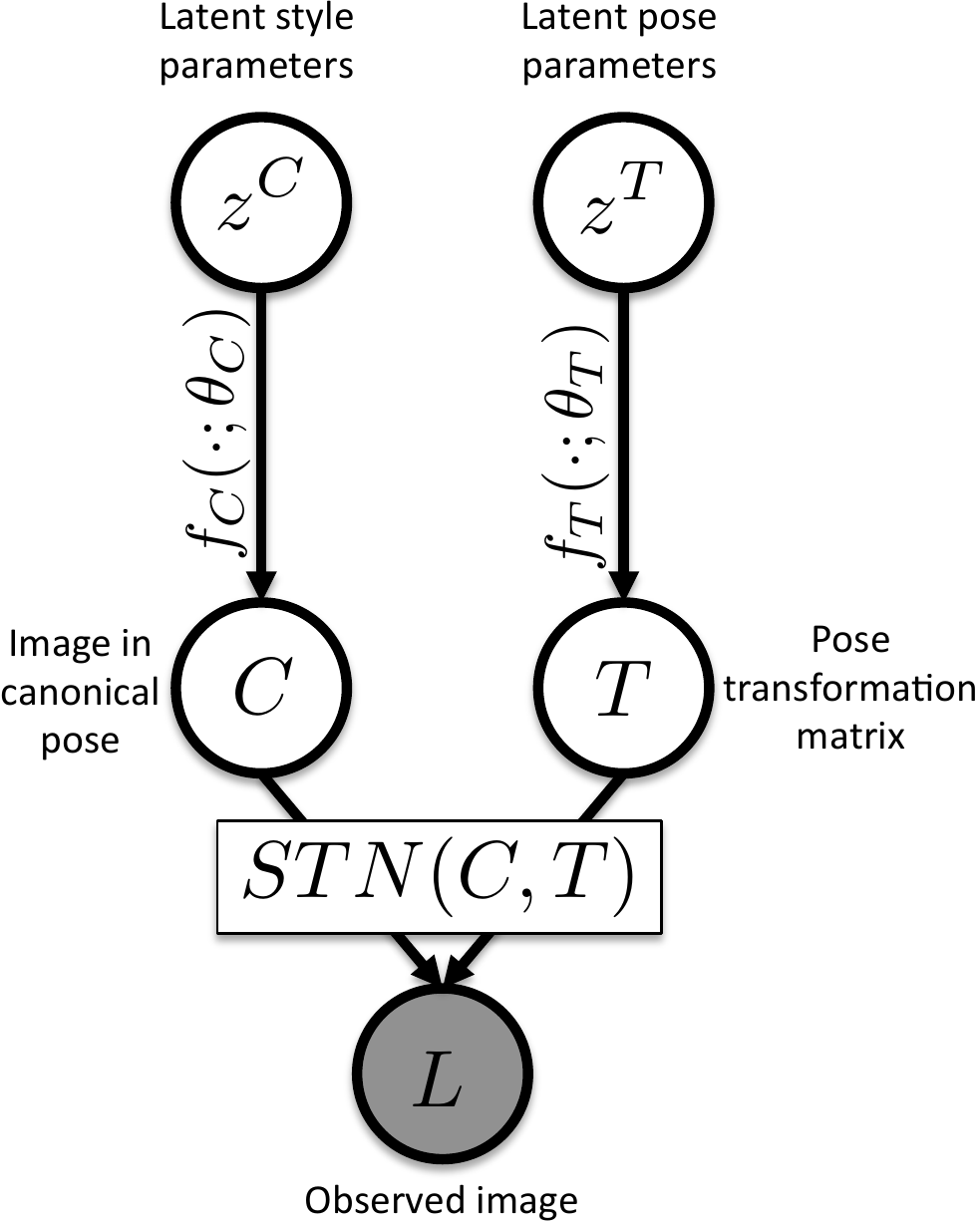}
\label{fig:stvae_decoder}
}\qquad\qquad
\subfigure[]{
\includegraphics[width=0.3\linewidth]{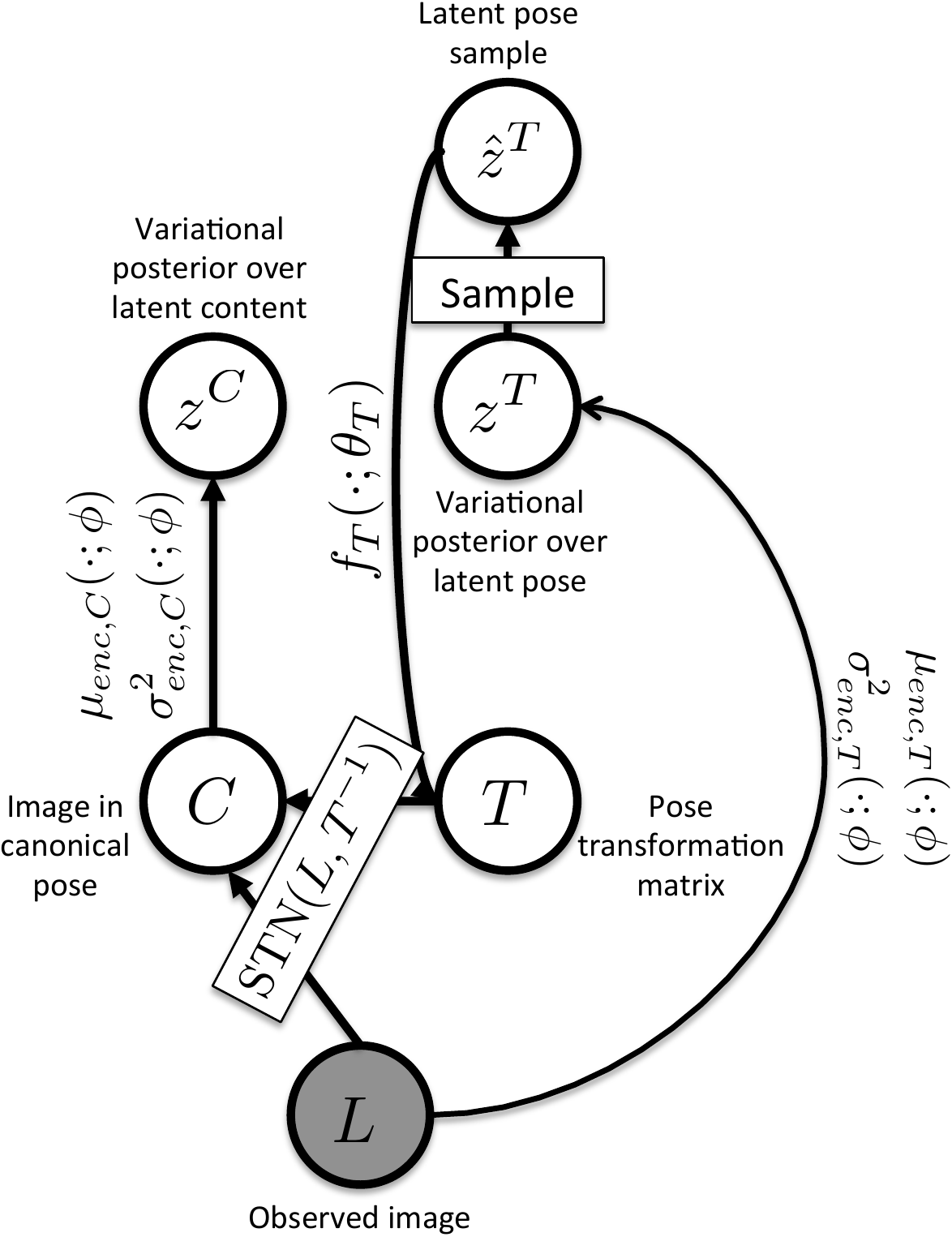}
\label{fig:stvae_encoder}
}\vspace{-5mm}
\end{center}
 \caption{\footnotesize
 \subref{fig:stvae_decoder} ST-VAE Generative model, $P(L | z^C, z^T)$ (Decoder); \subref{fig:stvae_encoder} ST-VAE Recognition model $Q(z^C,z^T | L) = Q(z^C | z^T, L)\cdot Q(z^T | L)$ (Encoder)   
  }\vspace{-5mm}
\label{fig:stvaemodel}
\end{figure}

We focus first on how to parameterize the recognition network (encoder) for the simpler case of a single layer model (i.e., the ST-VAE model
shown in Figure~\ref{fig:stvae_decoder}), in which we need only predict a single set of latent variables $z^C$ and $z^T$.
Na\"{i}vely, one could 
simply use an ordinary MLP to parameterize a distribution $Q(z^C, z^T | L)$, but ideally we would take advantage of the same
 insight that we used for the generative model,
namely 
 that it is easier to recognize content if we separately account for the
 pose.
To this end, we propose the ST-VAE recognition model shown in Figure~\ref{fig:stvae_encoder}. 
Conceptually the ST-VAE recognition model breaks the prediction of $z^C$ and $z^T$ into two stages.  Given the observed image $L$,
we first predict the latent representation of the pose, $z^T$.  Having this latent $z^T$ allows us to recover the pose transformation 
$T$ itself, which we use to ``undo'' the transformation of the generative process
by using the Spatial Transformer Network again but this time with the inverse transformation of the predicted pose.  This result, which can be
thought of as a prediction of the image in a canonical pose, is finally used to predict latent content parameters.

More precisely, we assume that the joint posterior distribution over
pose and content factors as
$Q(z^C, z^T | L) = Q(z^T |L)\cdot Q(z^C | z^T, L)$ where both factors are normal distributions.  
To obtain a draw $(\hat{z}^C, \hat{z}^T)$ from this posterior, we use the following procedure:\vspace{-4mm}

{\footnotesize
\begin{align*}
\hat{z}^T &\sim Q(z^T |L; \phi) = \mathcal{N}(\mu_{T}(L; \phi), \mbox{diag}(\sigma_{T}^2(L; \phi))), \\
\hat{T} &= f_T(\hat{z}^T; \theta_T), \\
\hat{C} &= \mbox{STN}(L, \hat{T}^{-1}), \\
\hat{z}^C &\sim Q(z^C | z^T, L; \phi) =  \mathcal{N}(\mu_{C}(\hat{C}; \phi), \mbox{diag}(\sigma_{C}^2(\hat{C}; \phi))),
\end{align*}\vspace{-4mm}
}

where $f_T$ is the pose decoder from the ST-VAE generative model discussed above.
To train an ST-VAE model, we then use the above parameterization of $Q$ and maximize Equation~\ref{eqn:vae_objective} with minibatch
SGD.
As long as the pose and content encoders and decoders  are differentiable, Equation~\ref{eqn:vae_objective} is guaranteed to also be 
end-to-end differentiable.

\subsection{Inference in the CST-VAE model}
\label{sec:cstvae}
\vspace{-3mm}

\begin{figure}[t]
\begin{center}
\includegraphics[width=0.65\linewidth]{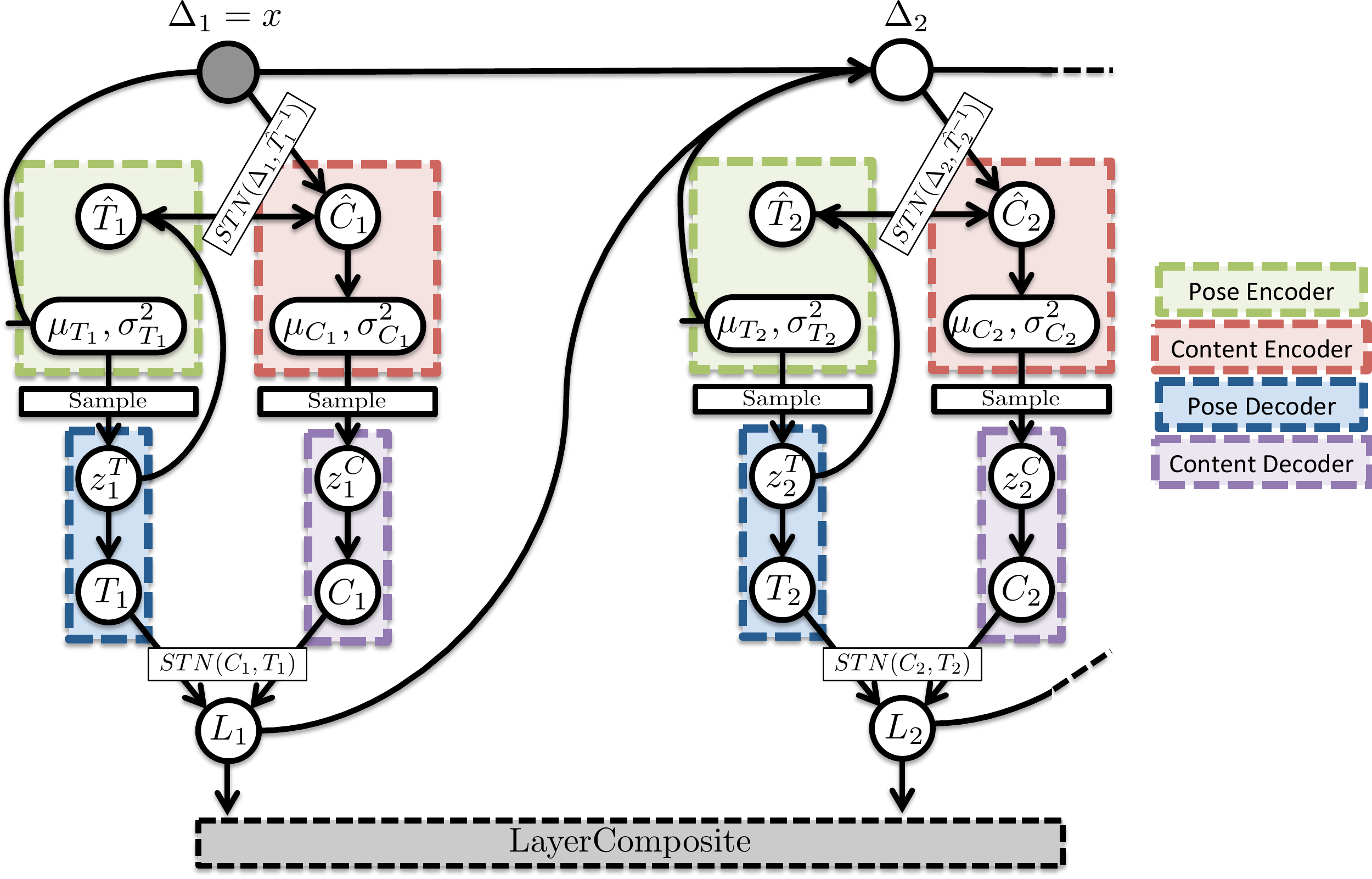}\vspace{-4mm}
\end{center}
 \caption{\footnotesize
The CST-VAE network ``unrolled'' for two image layers.
 }\vspace{-4mm}
\label{fig:cstvae}
\end{figure}

We now turn back to the multi-layer CST-VAE model, where again the task is to parameterize the recognition model $Q$. 
In particular we would like to 
avoid learning a model that must make a ``straight-shot'' joint prediction of all objects and their poses in an image.
Instead our approach is to perform inference over a single layer at a time from front to back, each time removing the contribution of a layer
from consideration until the last layer has been explained.

We proceed recursively: to perform inference for  layer $L_i$, we assume that the latent parameters $z^C_i$ and $z^T_i$ are responsible for explaining
some part of the residual image $\Delta_i$ --- i.e. the part of image that has not been explained by layers $L_1, \dots, L_{i-1}$ (note that $\Delta_1=x$).
We then use the ST-VAE module (both the decoder and encoder modules) 
to generate a reconstruction of the layer $L_i$ given the current residual image $\Delta_i$.  Finally to compute the next residual image to be explained by future layers, we set
$\Delta_{i+1} = \max (0, \Delta_i - L_i)$.  
We use 
the ReLU transfer function, $ReLU(\cdot)=\max(0, \cdot)$,  to ensure
that the residual image can always itself be interpreted as an image
(since 
$\Delta_i-L_i$ can be negative, which breaks interpretability of the
layers).

Note that our encoder for layer $L_i$ requires that the decoder has been run for layer $L_{i-1}$.  Thus it's not possible to separate the generative
and recognition models into disjoint parts as in the ST-VAE model.  Figure~\ref{fig:cstvae} unrolls the entire CST-VAE network (combining
both generative and recognition models) for two layers.
\vspace{-3mm}

\section{Evaluation}
\label{sec:eval}
\vspace{-2mm}

In all of our experiments we use the same training settings used in~\cite{Kingma2014}; that is,  
we use Adagrad for optimization with minibatches of 100  with a learning rate of 0.01
and a weight decay corresponding to a prior of $\mathcal{N}(0,1)$.
We initialize weights in our network using the heuristic of ~\cite{glorot2010understanding}.
However for the pose recognition modules in the ST-VAE model, we have found it useful to
specifically initialize biases so that poses are initially close to the identity transformation (see~\cite{jaderberg2015spatial}).

We use vanilla VAE models  as a baseline model against first the (single image layer) ST-VAE
model, then the more general CST-VAE model.  In all of our comparison we fix the training time for all models.
We experiment with between 20 and 50  dimensions for the latent content variables $z^C$
and always use 6 dimensions for pose variables $z^T$.
We parameterize content encoders and decoders
by using a two layer fully connected MLP with 256 dimensional
hidden layers and ReLU nonlinearities.
For pose decoders and encoders we also use two layer fully connected MLPs, but 
using 32 dimensional hidden layers and Tanh nonlinearities.
%
Finally for spatial transformer modules, we always resample onto a grid that is the same size as the original
image.


\begin{figure}[t]
\begin{center}
\subfigure[]{
\raisebox{2mm}{
\includegraphics[width=0.48\linewidth]{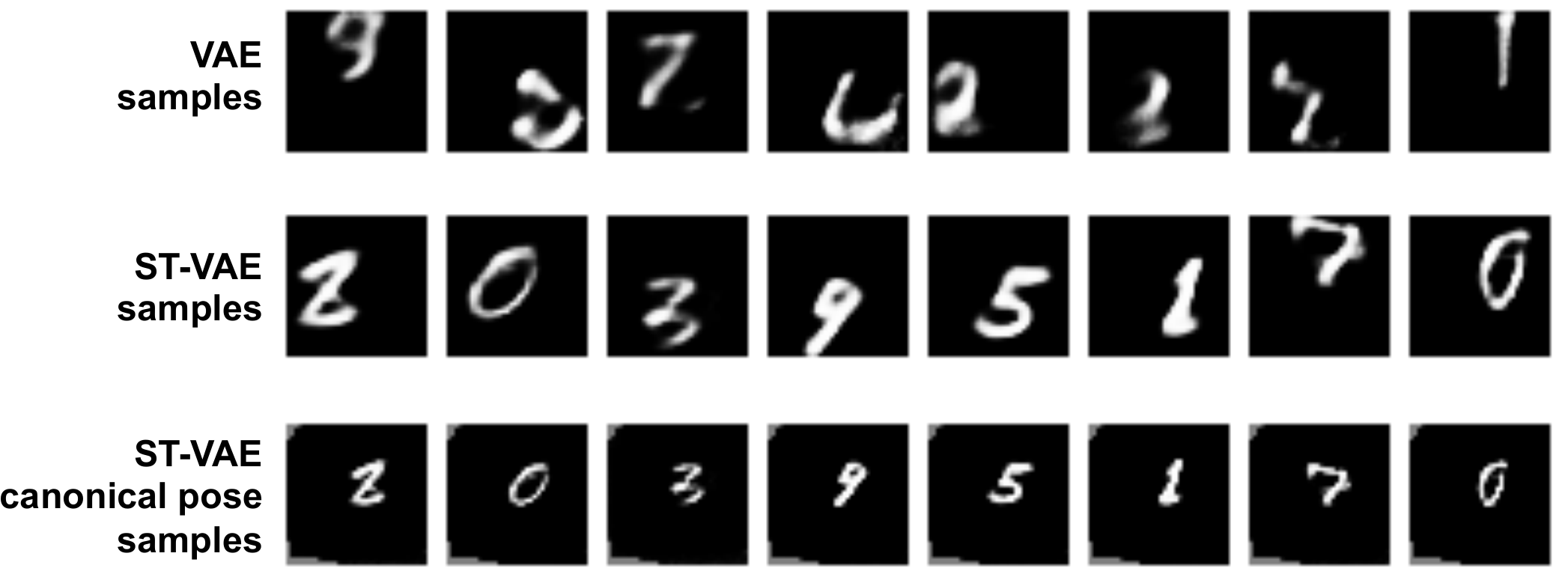}
\label{fig:vae_stvae_samples}
}
}
\,
\subfigure[]{
\includegraphics[width=0.45\linewidth]{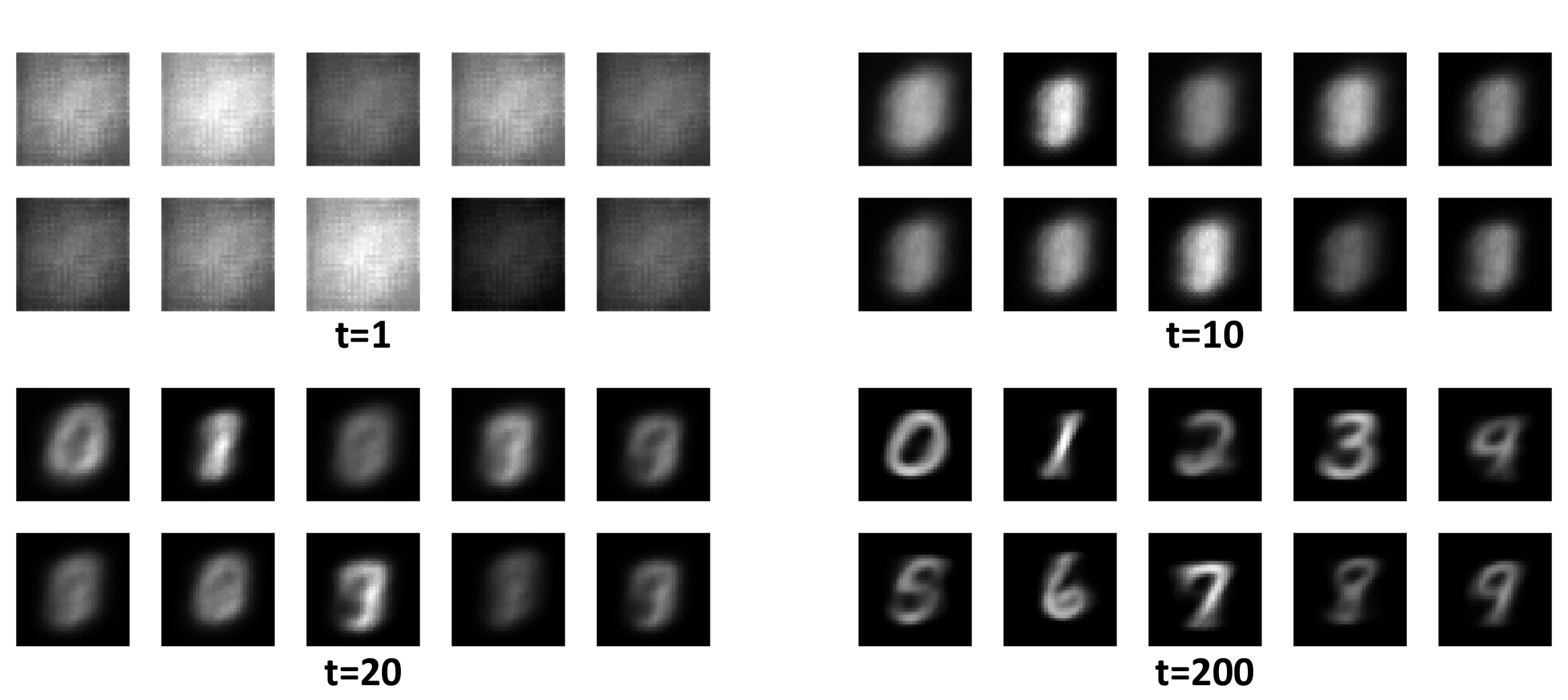}
\label{fig:stvae_averageddigits}
}\vspace{-5mm}
\end{center}
 \caption{\footnotesize
 \subref{fig:vae_stvae_samples} Comparison of samples from the VAE and ST-VAE generative models.  
 For the ST-VAE model, we show both the sample in its canonical pose and the final generated image.
  \subref{fig:stvae_averageddigits}  Averaged images from each MNIST class as learning progresses ---
  we typically see pose variables converge very quickly.
 }\vspace{-2mm}
\end{figure}

\begin{figure}[t]
\begin{center}
\subfigure[]{
\includegraphics[width=0.30\linewidth]{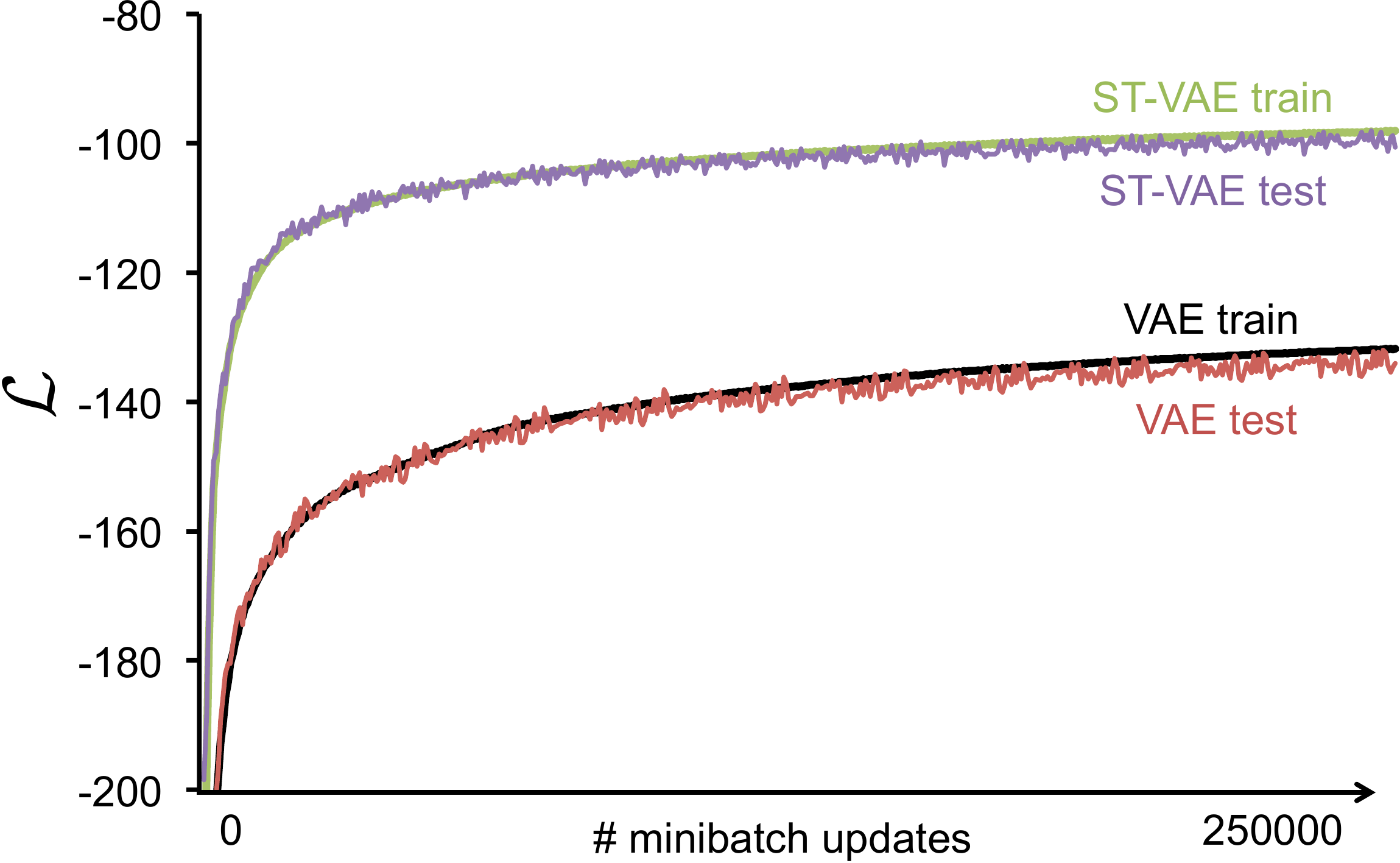}
\label{fig:vae_stvae_learning_curves}
}
\subfigure[]{
\raisebox{12mm}{
\footnotesize
\begin{tabular}{lcc} 
\hline 
 & Train Accuracy & Test Accuracy \\
\hline 
\multicolumn{3}{l}{On translated (36x36) MNIST} \\
\hline
VAE + supervised & 0.771 & 0.146 \\
ST-VAE + supervised & 0.972 & 0.964  \\
directly supervised & 0.884 & 0.783 \\
Directly supervised with STN & 0.993 & 0.969  \\
\hline 
\multicolumn{3}{l}{On original (28x28) MNIST} \\
\hline 
Directly supervised & 0.999 & 0.96   \\
\hline
\end{tabular}%
}
\label{fig:vae_stvae_classification}
}\vspace{-4mm}
\end{center}
 \caption{\footnotesize
 \subref{fig:vae_stvae_learning_curves} Train and test (per-example) lower bounds on log-likelihood 
for the vanilla VAE and ST-VAE models on the Translated MNIST data;
 \subref{fig:vae_stvae_classification} Classification accuracy obtained by supervised training using latent encodings
 from VAE and ST-VAE models.  More details in text.
 }\vspace{-2mm}
\end{figure}
\vspace{-2mm}
\subsection{Evaluating the ST-VAE on images of single objects}
\vspace{-2mm}

\looseness -1 We first evaluate our ST-VAE (single image layer) model alone on the MNIST dataset~\citep{lecun1998gradient}
and a derived dataset, \emph{TranslatedMNIST}, in which we randomly translated each  $28\times 28$ MNIST example
within a $36\times 36$ black image.  In both cases, we binarize the
images by thresholding,
as in~\cite{Kingma2014}.
Figure~\ref{fig:vae_stvae_learning_curves} plots train and test log-likelihoods over 250000 gradient steps
comparing the vanilla VAE model against the ST-VAE model, where we see that from the beginning the ST-VAE
model is able to achieve a much better likelihood while not overfitting.  This can also be seen in Figure~\ref{fig:vae_stvae_samples}
which visualizes samples from both generative models.   We see that while the VAE model (top row) manages to 
generate randomly transformed blobs on the image, these blobs typically only look somewhat like digits. 
For the ST-VAE model, we plot both the final samples (middle row) as well as the intermediate canonical images (last row), 
which typically are visually closer to MNIST digits. 

Interestingly, the canonical images tend to be slightly smaller versions
of the digits and our model relies on the Spatial Transformer Networks to scale them up at the end of the generative process.
Though we have not performed a careful investigation, 
possible reasons for this effect may be a combination of the fact (1) that scaling up the images introduces some blur which accounts for small variations in  nearby pixels and (2) it  is easier to encode smaller digits than larger ones.
 We also observe (Figure~\ref{fig:stvae_averageddigits})
that the pose network when trained on our dataset tends to converge rapidly, bringing digits to a centered canonical pose
within tens of gradient updates. 
Once the digits have been aligned, the content network is able to make better progress.

Finally we evaluate the latent content codes $z^C$ learned by our ST-VAE model 
in digit classification using the standard MNIST train/test split.  For this experiment
we use a two layer MLP with 32 hidden units in each layer and
ReLU nonlinearities
applied to the posterior mean of $z^C$ inferred from each image; we
do not use the labels to fine tune the VAE.
Figure~\ref{fig:vae_stvae_classification} summarizes the results, where we
compare against three baseline classifiers: (1) an MLP learned on latent codes from the VAE model,
(2) an MLP trained directly on Translated MNIST images (we call this the ``directly supervised classifier''), 
and (3) the approach of~\cite{jaderberg2015spatial} using the same MLP as above trained directly
on images but with a spatial transformer network.
As a point of reference, we also provide the performance of our classifier on the original $28\times 28$ MNIST
dataset.  We see that the ST-VAE model is able to learn a latent representation of image content that holds enough information
to be competitive with the ~\cite{jaderberg2015spatial} approach (both of which slightly outperform 
the MLP training directly on the original MNIST set).  The approaches that do not account for
pose variation do much worse than ST-VAE on this task and exhibit significant overfitting.

\vspace{-2mm}
\subsection{Evaluating the CST-VAE on images with multiple overlapping
  objects}
\vspace{-2mm}

\begin{figure}[t]
\begin{center}
\subfigure[]{
\raisebox{7mm}{
\includegraphics[width=0.35\linewidth]{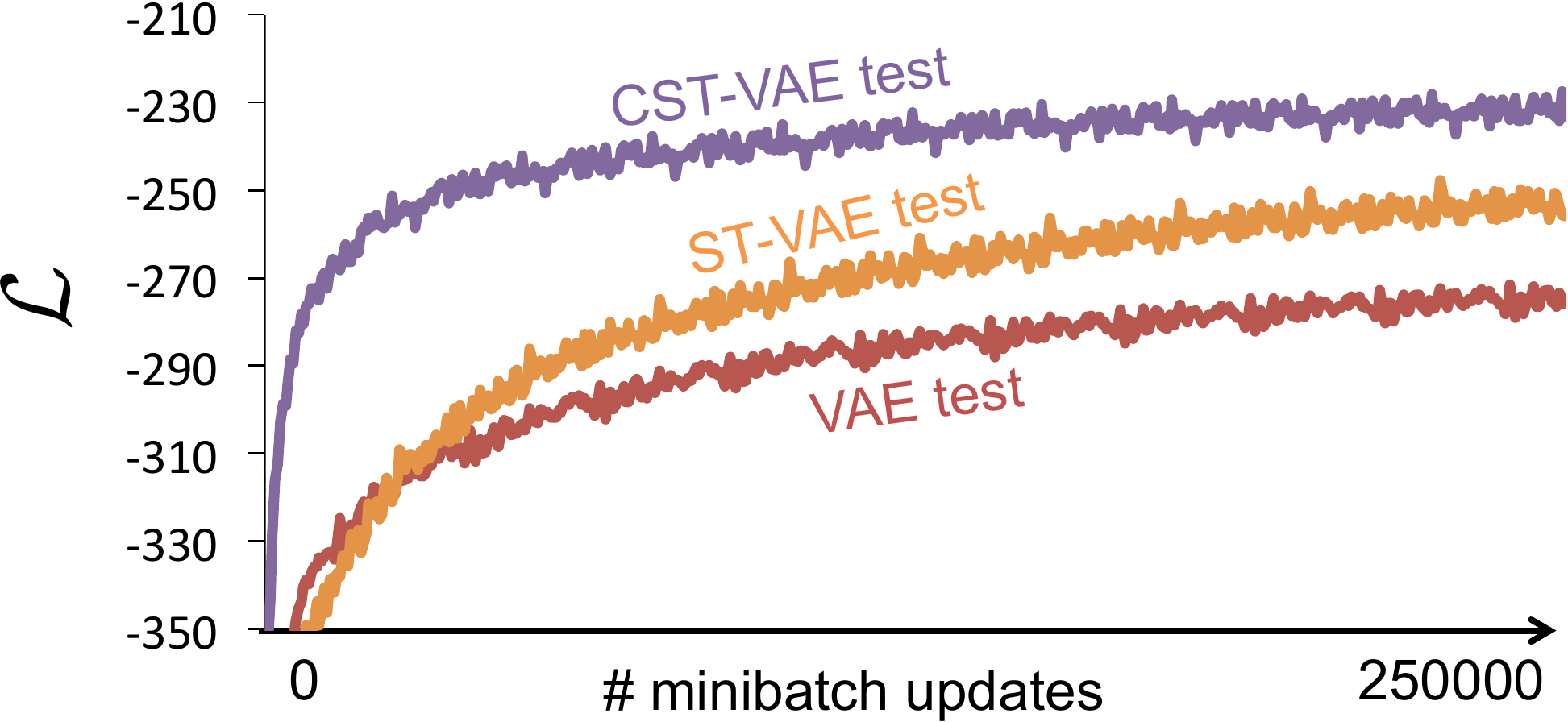}
\label{fig:cstvae_learning_curves}
}
}\;
\subfigure[]{
\raisebox{5mm}{
\includegraphics[width=0.2\linewidth]{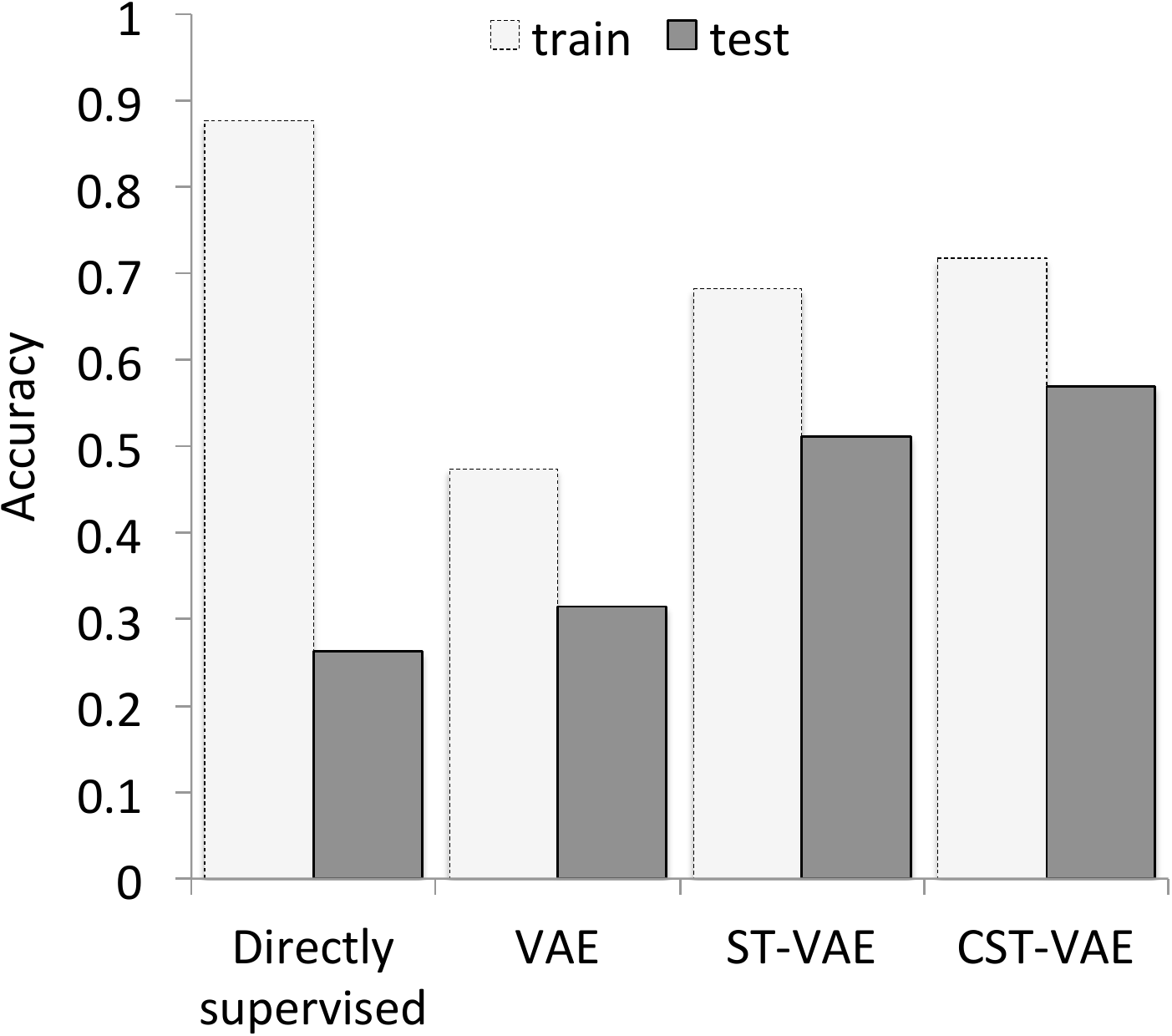}
\label{fig:cstvae_classification}
}
}\;
\subfigure[]{
\includegraphics[width=0.35\linewidth]{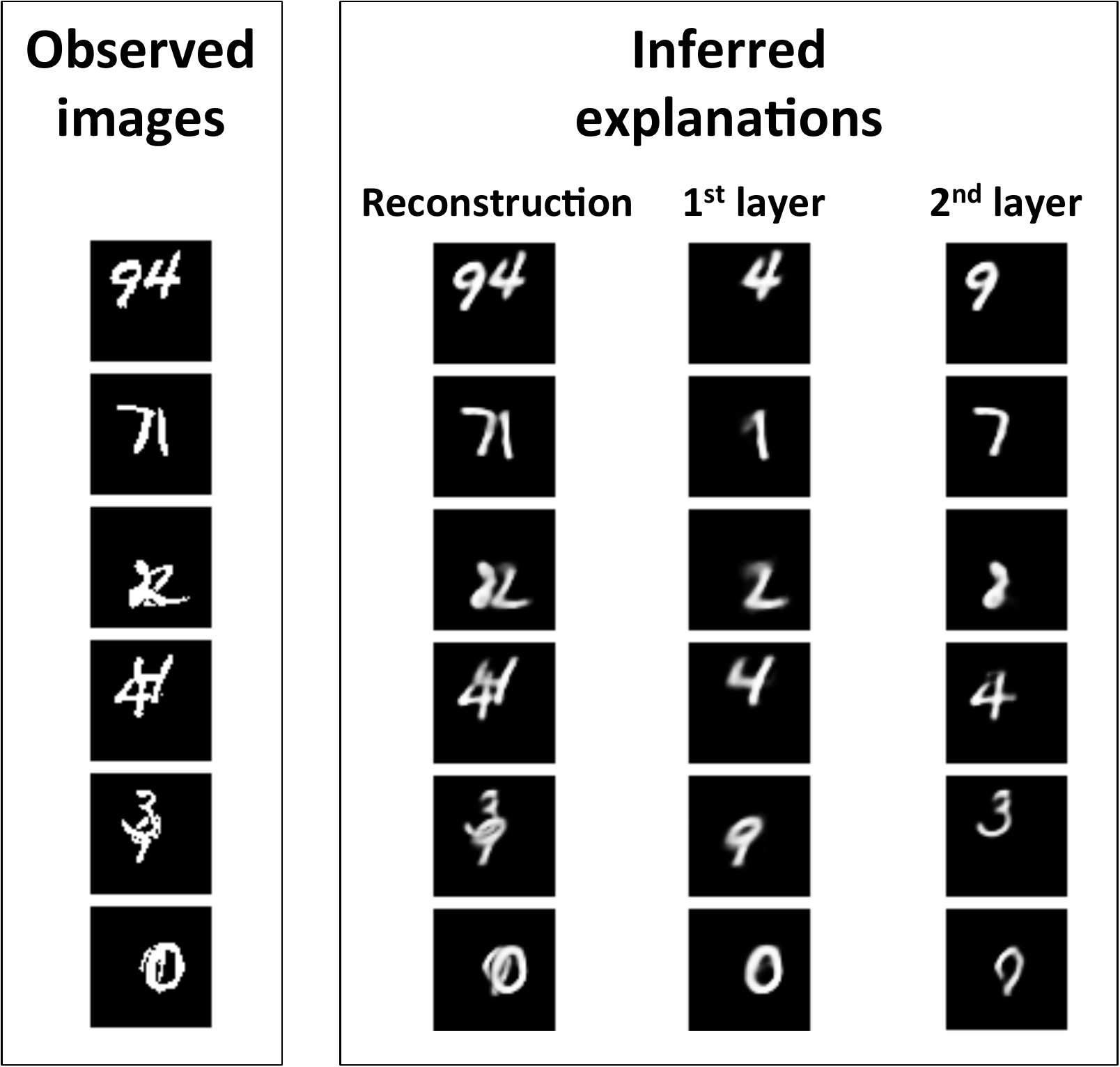}
\label{fig:cstvae_reconstructions}
}\vspace{-5mm}
\end{center}
 \caption{\footnotesize
 \subref{fig:cstvae_learning_curves} Train and test (per-example) lower bounds on log-likelihood 
for the vanilla VAE and CST-VAE models on the Superimposed MNIST data;
  \subref{fig:cstvae_classification}
 Classification accuracy obtained by supervised training using latent encodings
 from VAE and CST-VAE models.
  More details in text.
 \subref{fig:cstvae_reconstructions}  Images from the Superimposed MNIST dataset with visualizations of
 intermediate variables in the neural network corresponding to first and second image layers and the final reconstruction.
}\vspace{-3mm}
\end{figure}


We now show results from the CST-VAE model on a challenging ``Superimposed MNIST'' dataset.
We constructed this dataset
by randomly translating then superimposing two MNIST digits one at a time onto $50\times 50$ black backgrounds,
generating a total of 100,000 images for training and 50,000 for testing.  
A large fraction of the dataset thus consists of overlapping digits that occlude one another, sometimes so severely that
a human is unable to classify the two digits in the image.
In this section we use the same pose/content encoder and decoder
architectures as above except that we set hidden content encoder/decoder layers to be 128-dimensional --- empirically,
we find that larger hidden layers tend to be sensitive to initialization for this model.
We also assume that observed images are composited
using two image layers (which can be thought of as foreground and background).

Figure~\ref{fig:cstvae_learning_curves}
plots test log-likelihoods over 250000 gradient steps
comparing the vanilla VAE model against the ST-VAE and CST-VAE model, where 
we see that from the beginning the CST-VAE
model is able to achieve a much better solution than the ST-VAE model
which in turn outperforms the VAE model.   
In this experiment, we ensure that the total number of latent dimensions across all
models is similar.
In particular, 
we allow the VAE and ST-VAE models to use 50 latent dimensions for content.
The ST-VAE model uses an additional 6 dimensions for the latent pose.
For the CST-VAE model we use 20 latent content dimensions
and 6 latent pose dimensions per image layer $L_i$ (for a total of 52 latent dimensions).

Figure~\ref{fig:cstvae_reconstructions} highlights the interpretability of our model.
On the left column, we show example superimposed digits from our dataset and ask the 
CST-VAE to reconstruct them (second column).  As a byproduct of this reconstruction,
we are able to individually separate a foreground image (third column) and background image (fourth column),
often corresponding to the correct digits that were used to generate the observation.  
 While not perfect, the CST-VAE model manages to do well even on some challenging examples where
 digits exhibit high occlusion.  To generate these foreground/background images, we use the posterior mean
 inferred by the network for each image layer $L_i$; however, we note that one of the advantages of the 
 variational auto-encoder framework is that it is also able to represent uncertainty over different interpretations
 of the input image.

Finally, we evaluate our performance on classification.
We use the same two layer MLP architecture (with 256 hidden layer units) as we did 
with the ST-VAE model, and train using latent representations
learned by the CST-VAE model.  
Specifically we concatenate the latent content vectors $z^C_1$ and $z^C_2$
which are fed as input to the classifier network.  
As baselines we compare against (1) the vanilla VAE latent representations 
and (2) a classifier trained directly on images of superimposed
digits.  We report accuracy, requiring that the classifier 
be correct on both digits within an image.\footnote{
Thus  
 chance performance on this task is 0.018 (1.8\% accuracy) since we
 require that the image recover both digits correctly 
 within an image.
}

Figure~\ref{fig:cstvae_classification} visualizes the results.  
We see that the classifier that is trained directly
on pixels exhibits severe overfitting and performs the worst.  The three variational auto-encoder models also
slightly overfit, but perform better, with the CST-VAE obtaining the
best results, with almost twice the accuracy as the vanilla 
VAE model.

\vspace{-2mm}



\section{Conclusion}
\vspace{-2mm}

We have shown how to combine an old idea --- of interpretable, generative, layered models of images --- 
with modern techniques of deep learning, in order to tackle the challenging problem of intepreting images in the presence of occlusion in an entirely unsupervised fashion. We see this is as a crucial stepping stone to future work on deeper scene understanding, going beyond simple 
feedforward supervised prediction problems.
In the future, we would like to apply our approach to real images, and possibly video.
This will require extending our methods to use convolutional networks, and may
also require some weak supervision (e.g., in the form of observed object class labels associated with layers)
or curriculum learning to simplify the learning task.

\eat{
Probabilistic graphical models with generative semantics were popular in vision not too many years ago 
but in recent years have largely fallen out of favor. These more traditional generative models typically are burdened by slow inference and can be surprisingly unwieldy since good accuracy often relied on maintaining code for a large family of different handcrafted visual features which had to be tuned appropriately.  However our work suggests that we may not want to throw the baby out with the bathwater just yet.  
Probabilistic generative models often have interpretable  semantics  and allow us to learn with supervision, weak/semi-supervision and no supervision all within a unified framework.  Our models are examples of this type of interpretable generative model that \emph{can} coexist with rich feature hierarchies that are automatically tuned by end-to-end backpropagation. They are also fast at test time, requiring just a single forward pass through a neural network.  Finally they can be trained with no labeled data. Thus we believe that this combining of the best of both worlds will be a fruitful area of future work in both representation learning
and computer vision.
}


\subsubsection*{Acknowledgments}
We are grateful to Sergio Guadarrama and Rahul Sukthankar for reading and providing feedback on 
a draft of this paper.

\bibliography{iclr2016_conference}
\bibliographystyle{iclr2016_conference}

\end{document}